# Action parsing using context features

Nagita Mehrseresht
Canon Information Systems Research Australia,
5 Talavera rd., North Ryde, Australia

*Abstract*—

We propose an action parsing algorithm to parse a video sequence containing an unknown number of actions into its action segments. We argue that context information, particularly the temporal information about other actions in the video sequence, is valuable for action segmentation. The proposed parsing algorithm temporally segments the video sequence into action segments. The optimal temporal segmentation is found using a dynamic programming search algorithm that optimizes the overall classification confidence score. The classification score of each segment is determined using local features calculated from that segment as well as context features calculated from other candidate action segments of the sequence. Experimental results on the Breakfast activity data-set showed improved segmentation accuracy compared to existing state-of-the-art parsing techniques.

## I. INTRODUCTION

Understanding human activities in videos is a challenging problem that has been studied in the context of many applications such as surveillance, healthcare, human computer interactions, robot design, sport analytics, video summarization, and automatic content based video annotation and retrieval.

Despite extensive research in human activity recognition for over 20 years, the problem of understanding complex human activities in complex scenes is generally unsolved. Human action recognition is particularly challenging due to self-occlusion, interaction with objects, variation in different instances of a same action, and visually similar but different actions. These challenges are besides common challenges of scene analysis in computer vision, such as background clutter, variation in illumination, scale, orientation and view-point.

The majority of existing human activity recognition research focuses on action classification, which is the problem of assigning a single class to a predetermined video segment. Experiments on common data-sets such as KTH, Weizmann, Hollywood2, UCF50 and UCF101 generally report such classification results [4, 11, 14, 16, 19]. Such techniques assume that the video sequence can be pre-segmented into coherent segments containing a single action of interest and minimal background content. The pre-segmentation is often done manually and the challenging problem of finding the temporal segment which contains the action is ignored.

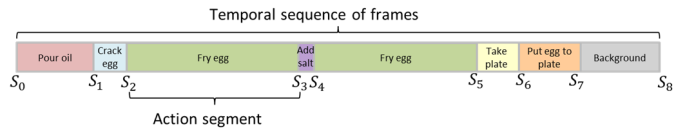

Fig. 1. Temporal parsing of a sequence $(X_{0:N})$ to action segments.

In this paper, we use the term *action segment* to refer to the temporal segment that contains an instance of an action of interest. Action localization, which may also be referred to as temporal segmentation, is the task of finding the action segment of an action of interest; thus it includes the two sub-tasks of localizing the action segment by finding the start and the end frames of the segment, and finding the action classification label associated with that segment. A temporal segment of a video is a continuous set of frames from frame $f_1$ to frame $f_2$, where frame $f_1$ is temporally before frame $f_2$; i.e., $f_1 \leq f_2$. Fig. [1] shows the action segments in an exemplary temporal sequence of frames.

For human action recognition to be generally useful in practical applications, the action localization problem cannot be ignored. Action localization, however, is a challenging problem since different instances of a same action could have different durations, and there could be more than an order of magnitude difference between the temporal duration of different actions of interest. For example, while repetitive actions like walking and running may last for a few seconds to many seconds, snap actions like kicking, falling and kissing last for a fraction of a second to a few seconds. Also, without knowing the temporal localization, a segment of one action, for example a stride in a walking action may look like a different action, e.g., kicking.

A sliding window search approach is commonly used for temporal action segmentation [3, 7, 11, 12, 21, 25, 26] and includes training a discriminative model for the actions of interest, using a training set containing segmented instances of the actions of interest. The trained model is then applied to a set of fixed length and often overlapping temporal segments of an unseen video. The length of the segments and the ratio of the overlap are generally predetermined, and using different lengths, e.g., multi-scale segmentation, is common. The final action segments are then determined by non-max suppression which greedily selects the segments with the highest scores [12, 15]. A disadvantage of using a sliding window search is that the precision of localization depends on the resolution of the search and therefore the number of evaluated temporal segments. Also as the final segmentation is done locally and using greedy algorithm, the generated temporal segments are not jointly optimized.

We argue that context information—particularly the temporal information about the long-term activity that the action is part of and the other actions which are also part of that long term activity—is beneficial for action localization. Action classification accuracy can be improved if in addition to visual information of each segment we also use classification information of other segments of the video as context features. Using context feature for action localization, however, is more challenging than using context features for classification, as it would require access to classification information of other action segments which are also not yet known at the time of finding all action segments jointly.

In this work we propose a new temporal action segmentation technique which jointly performs temporal segmentation and classification and generates a parsing of a temporal sequence of frames into its action segments. The temporal segmentation is globally optimized over the entire given sequence of frames. One of the contributions of this work is a new action parsing technique, which uses temporal context features generated from other segments to improve classification and localization accuracy.

The following section contains a literature review of other works relevant to action classification and localization. Details of the proposed action parsing algorithm and the used temporal context features are given in sections III-A and III-B respectively. Section IV includes the details of the evaluation methodology and results are shown in section V. Conclusions are provided in Section VI.

## II. RELATED WORK

Common action classification techniques use a bag-of-words [24] or Fisher vector encoding [25] of local space-time features and a discriminative classifier, such as support vector machine (SVM), to assign a class label to an input video segment. Earlier space-time features include space-time interest points (STIP) [9] and 3D extensions of common image descriptors such as 3D HOG [6], 3D-SIFT [17] and extended SURF [28]. More recently improved dense trajectories (iDT) have shown promising performance for video action classification [25]. Action classification using features learned by convolutional neural networks (CNNs), such as two-stream convolutional networks [18] and the C3D method of [20], has also shown better or comparable classification accuracy to traditional features.

A common approach to finding the temporal segments containing an action of interest is to integrate discriminative classification with a sliding window search and non-max suppression. Laptev et al. performed a sliding window search on cuboids, thus restricting the action to have a fixed spatial extent across frames [10]. [3] proposes an actom sequence model and performs temporal action localization using a sliding central frame approach. [7] proposes a track aligned 3D-HOG action representation and detects specific actions within tracks using a sliding window classification approach. [26] first detects action proposals at the frame-level and scores them with a combination of static and motion CNN features. Temporal localization of actions is performed using a sliding-window approach at the track level. [21] generalizes the deformable part models from 2D images [2] to 3D spatiotemporal volumes, and uses a sliding window approach in scale, space and time to detect actions in videos. [25] uses Fisher vector encoding of iDT features and localizes the actions using a sliding window search. To speed up the localization search Oneata et al. proposed an approximately normalized Fisher Vector encoding technique [12].

Existing techniques for joint segmentation and classification are often based on generative models. [27] uses stochastic grammar and a Bayesian network called a sequential interval network (which models the start and the end times of actions) to parse a sequence as a series of (sub-)actions. Kuehne et al. modelled complex activities as temporally structured processes using a combination of a context free grammar (CFG) and a hidden Markov model (HMM) [8]. Their approach used HMMs to recognize actions. Pirsiavash et al. used a context-free grammar and a latent structural SVM [13]. A limitation of grammar-based approaches is that they are only efficient when the actions of interest can be described as a combination of a limited number of sub-actions.

Hoai et al. proposed a discriminative joint temporal segmentation and classification of actions in video using a dynamic programming search to inference the temporal segmentation [5]. However, in the method proposed in [5] the classification of each action segment is performed using features extracted from frames in that segment only. Our proposed action parsing technique addresses a shortcoming of [5], by incorporating temporal context. We demonstrated that the use of temporal context features improves the action parsing accuracy.

## III. PROPOSED ACTION SEQUENCE PARSING METHOD

This section describes our proposed action parsing algorithm which can be used to parse a video sequence into its action segments. The proposed algorithm optimizes the sum of classification confidence scores of all selected action segments. The classification score of each segment is determined using local features calculated from that segment as well as context features calculated from other parts of the sequence. Parsing is performed using a dynamic programming search algorithm which is an extension of the method proposed in [5].

### A. Parsing algorithm

As shown in Fig. [1], for a given temporal sequence $X_{0:N}$ of length $N + 1$ frames, the goal of temporal parsing is to find the optimal set of breakpoints $S_o, S_1, \ldots, S_K$ which segments the sequence into action segments, where each action segment contains all frames of (an instance of) an action of interest. $K$ is the number of action segments in the sequence $X_{0:N}$ and is not known in advance. The value of $K$ is determined as part of

```
Algorithm:  Forward pass; finding the optimal parsing
scores
Initialization: $\gamma \leftarrow [-\infty]_N, \beta \leftarrow [0]_N, \rho \leftarrow [0]_N$
Repeat:
    For $u = 1:N$
        For $l = l_{min}:l_{max}$
            If $\xi(u,l) + \gamma(u-l) > \gamma(u)$ do
                $\gamma(u) \leftarrow \xi(u,l) + \gamma(u-l)$
                $\beta(u) \leftarrow c_{u-l:u}$
                $\rho(u) \leftarrow l$
```

Fig. 2. Forward pass of temporal parsing optimization algorithm

```
Algorithm:  Backward pass; finding segment lengths and
class labels of the temporal sequence $X_{0:N}$
Initialization: $i \leftarrow N, j \leftarrow 1$
Repeat:
    While $i \geq 0$ do
        $C(j) \leftarrow \beta(i)$
        $S(j) \leftarrow i$
        $i \leftarrow i - \rho(i)$
        $j \leftarrow j + 1$
Do
Reverse the order of elements in $C$ and $S$.
```

Fig. 3. Backward pass of the temporal parsing algorithm.

the optimization problem which needs to be solved for each given sequence.
The set of breakpoints $S_o, S_1, \ldots, S_K$ must provide a temporal parsing of the temporal sequence $X_{0:N}$. In particular, the segmentation algorithm must satisfy the following additional constraints:
- "No-gap" requirement: ; i.e., the unions of all action segments must equal the original sequence:
$$\bigcup_{i=0 \ldots K} X_{S_i:S_{i+1}} = X_{0:N}$$
- The action segments must not overlap:
$$\forall i,j, i \neq j \quad X_{S_i:S_{i+1}} \cap X_{S_j:S_{j+1}} = \emptyset$$
- Each segment must refer to a set of consecutive frames.

The parsing algorithm may also need to satisfy constraints on the length of the segments; e.g.,
$$l_{min} \leq len(X_{S_i:S_{i+1}}) \leq l_{max}.$$
$l_{min}$ and $l_{max}$ respectively correspond to the minimum and the maximum segment lengths that would be considered by the parsing algorithm.
To satisfy the no-gap requirement, a special 'background' class is often added to the list of actions of interest to cover the intervals where no action of interest happens. In particular, the class 'background' includes both idle and any other action which is not of interest. For the purpose of temporal parsing, 'background' class is just like another action class of interest, and will not be mentioned as a special class in the rest of this work.
Given a pre-trained classification model, a scoring function, $\mathcal{F}_{score}$ can be formed which given a segment $X_{S_i:S_{i+1}}$ returns a classification label $c_i$ of the class of the action in the segment, and a confidence score $d_i$ ($d_i \geq 0$) of classifying the segment to the class $c_i$; i.e.,
$$\mathcal{F}_{score}: X_{S_i:S_{i+1}} \rightarrow (d_i, c_i) \quad (1)$$
The scoring function $\mathcal{F}_{score}$ will be discussed in details in section III-A.
At recall time the optimal set of breakpoints $S_o, S_1, \ldots, S_K$ are found by maximizing the overall confidence score, $\sum_{i=0}^{K} d_i$;
$$\max_{K, S_0, \ldots, S_K} \sum_{i=0}^{K} d_i \quad (2)$$

Hoai et al. proposed a dynamic programming algorithm which can be used to solve optimization problems similar to Eq. (2) [5].
The dynamic programming algorithm solves the optimization problem of Eq. (2) by finding the optimal solution for shorter sub-sequences. For a given sub-sequence $X_{0:u}$ ending at point $u$, $u = 1, \ldots, N$, we use $\gamma(u)$ to represent the optimal parsing score for the sub-sequence $X_{0:u}$.
For every tuple $(u, l)$, $u \in \{1, \ldots, N\}$ and $l \in \{l_{min}, \ldots, l_{max}\}$
$$\xi(u,l) = d_{u-l:u};$$
represents the confidence score of classifying the candidate segment $X_{u-l:l}$ to one of the actions of interest. As proposed in [5] $\gamma(u)$ can be efficiently calculated using dynamic programming by finding the optimal segment length $l$ when solving the optimization problem of
$$\gamma(u) = \max_{l_{min} \leq l \leq l_{max}} \xi(u,l) + \gamma(u-l) \quad (3)$$
Fig. 2 is an algorithm which can be used to solve the optimization problem of Eq. (3). In this algorithm, $[0]_N$ and $[-\infty]_N$ represents vectors of length $N$ initialized to zero and $-\infty$ respectively.

Using the parsing algorithm of Fig. [2], $\gamma$ contains the optimal parsing score for each sub-sequence $X_{0:u}$ ending at point $u$, $u \in \{1, \ldots, N\}$; $\beta$ and $\rho$ contain information about the class labels and segments' length respectively.
Once the values of $\gamma$, $\beta$ and $\rho$ are determined for all end-points $u$, the optimal parsing and the set of class labels can be found by using the backward pass as shown in Fig. [3].

Using the backward pass of Fig. [3], $K = len(S) + 1$; S is the set of segment breakpoints $S_o, \ldots, S_K$ and $C$ contains the corresponding classification labels of each estimated action segment.

### B. Action classification using context features

Accuracy of temporal segmentation and the corresponding sequence of classification labels generated using the parsing method of section III-A depends on the accuracy of the scoring function $\mathcal{F}_{score}$ of Eq. (1).

One problem with finding the optimal segmentation using a classifier which evaluates each segment individually is that when the classification task is hard, the classification confidence generally is low and too noisy to accurately guide the temporal segmentation. When the input sequence containing multiple action segments, the recognition accuracy often improves significantly when instead of classifying each segment individually, we classify the segments jointly, for example using a structured model such as conditional Markov random field (CRF) [22] or HMM [23,29]. Standard discriminative joint classification models such as CRF, however, cannot be used with the parsing algorithm of section III-A as a joint classifier would require all segments to be defined a priori.

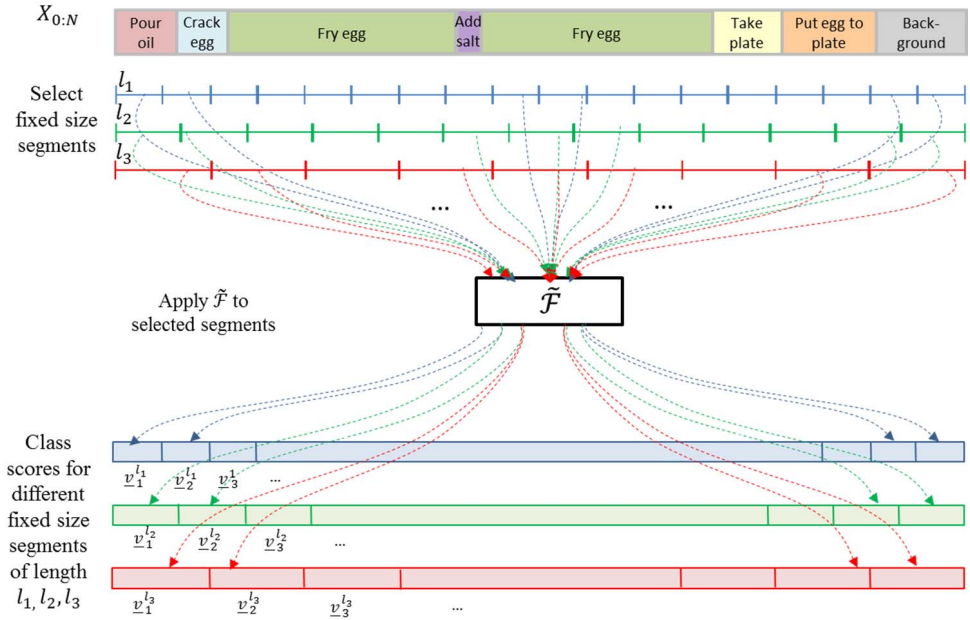

Fig. 4 Generating temporal context features by calculating initial classification scores for fixed size segments.

Temporal context information, such as the knowledge of what other actions have happened before or after a segment of interest can also improve the recognition accuracy. However, determining what other actions have happened would require knowing the temporal segmentation before and after the segment of interest. The temporal segmentation information is not available when the parsing algorithm is jointly searching for the best overall segmentation of the input sequence.

In this work, we propose to use multi-scale fixed size segmentation and max pooling to generate temporal context features for an improved action classifier which can efficiently be used in the multi-segment scoring function $\mathcal{F}_{\text{score}}$. The proposed improved action classification technique uses two discriminative classifiers.

The first layer classifier, used for generating context features, is a multi-class support vector machine (SVM) trained using a collection of training segments $X^1, \ldots, X^n$ each containing an action of interest, or being an instance of the background class, and the corresponding class labels. This classifier is applied to each segment independently and uses an encoding of the local features in segments $X^i$ denoted as $\varphi(X^i)$. The parameters $\widetilde{\omega}$ of such classifier are learned by optimizing

$$\underset{\widetilde{\omega}_j, \vartheta^i \geq 0}{\text{minimize}} \frac{1}{2m}\sum_{j=1}^{m}\|\widetilde{\omega}_j\|^2 + \frac{1}{\lambda}\sum_{i=1}^{n}\vartheta^i \quad s.t. \quad (4)$$

$$(\widetilde{\omega}_{y^i} - \widetilde{\omega}_y)^T \varphi(X^i) \geq 1 - \vartheta^i, \forall i, y \neq y^i$$

$y^i$ is the ground truth class label associated with the segment $X^i$. An example of $\varphi(X^i)$ is a Fisher vector encoding of STIP [9] or iDT [24] features calculated from $X^i$. Here, $\widetilde{\omega}_y^T \varphi(X^i)$ is the SVM score for assigning $X^i$ to class $y$.

The SVM trained as above is used as temporal context scoring function $\widetilde{\mathcal{F}}$. As shown in Fig. [4], given a segment $\widetilde{X}$ of length $l_i$, the temporal context scoring function $\widetilde{\mathcal{F}}$ returns a vector $v^{l_i}$; the $j^{th}$ element of $v^{l_i}$ is the SVM score for classifying $\widetilde{X}$ as the $j^{th}$ class.

$$\widetilde{\mathcal{F}}: \widetilde{X}_{t:t+l_i} \to v^{l_i} \quad (5)$$
$$v^{l_i}[j] = \omega_j^T \cdot \varphi(\widetilde{X}_{t:t+l_i})$$

Alternatively, an action classifier using CNN and softmax layer such as [18, 20] can be used as temporal context scoring function $\widetilde{\mathcal{F}}$. In that case, the temporal context vector $v^{l_i}$ is the softmax score for different action classes.

As shown in Fig. [4], to generate temporal context features for a given sequence, we first use multi-scale fixed size segmentation and apply the context scoring function $\widetilde{\mathcal{F}}$ to all segments of length $l_i$; $l_i \in \{l_1, l_2, \ldots, l_w\}$, where $w$ is a predefined number of scales. The vectors $\underline{v}_j^{l_i}$ determined for all segments $j$ of length $l_i, j \in \{1, \ldots, \lceil\frac{N+1}{l_i}\rceil\}$, are cached for further processing which would generate context features by max-pooling subsets of $\underline{v}_j^{l_i}$ values.

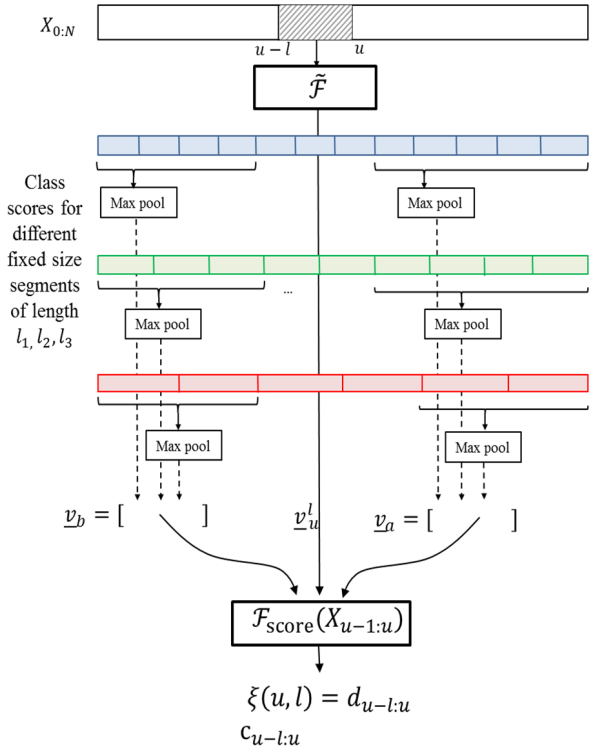

Fig. 5. Proposed scoring function which uses temporal context features.

Fig. [5] illustrates the proposed multi-segment scoring function $\mathcal{F}_{score}$ which is used to determine the confidence score $\xi(u, l)$ in the parsing algorithm of section III-A. As shown in Fig. [5], for any candidate segment $X_{u-l:u}$ in the forward pass of Fig. [2], we determine $\tilde{\mathcal{F}}(X_{u-l:u}) = \underline{v}_u^l$. Separately at each segmentation scale $l_i$, we apply per-class max-pooling of $\underline{v}_j^{l_i}$ scores, of all segments before ($j \leq \left\lfloor \frac{u-l}{l_i} \right\rfloor$) and after ($j \geq \left\lceil \frac{u}{l_i} \right\rceil$) the segment $X_{u-l:u}$. The concatenation of the max-pooled scores of all scales $l \in \{l_1, l_2, ..., l_w\}$ occurring before the segment $X_{u-l:u}$ is denoted as $\underline{v}_b$. Similarly, the concatenation of the max-pooled scores of all scales $l \in \{l_1, l_2, ..., l_w\}$ occurring after the segment $X_{u-l:u}$ is called $\underline{v}_a$.

$\mathcal{F}_{score}$ determines the classification label $c_{u-l:u}$ and the classification confidence score $\xi(u, l)$ of the segment $X_{u-l:u}$ by applying a second layer classifier which uses the concatenation of $\underline{v}_b$, $\underline{v}_u^l$ and $\underline{v}_a$ denoted as

$$\psi(X_{u-l:u}) = \begin{bmatrix} \underline{v}_b, & \underline{v}_u^l, & \underline{v}_a \end{bmatrix}$$

as the input feature vector. The second layer classifier is also a multi-class SVM and is also trained using the collection of training segments $X^1, ..., X^n$ and the corresponding classification labels, but using the concatenated scores $\psi(X_{u-l:u})$. The parameters $\omega_j$ of the second layer classifier are determined by optimizing

$$\underset{\omega_j,\, \vartheta^i \geq 0}{\text{minimize}} \frac{1}{2m} \sum_{j=1}^{m} \|\omega_j\|^2 + \frac{1}{\lambda} \sum_{i=1}^{n} \vartheta^i \quad s.t. \quad (6)$$

$$(\omega_{y^i} - \omega_y)^T \psi(X^i) \geq 1 - \vartheta^i, \forall i, y \neq y^i$$

IV. EVALUATION METHODOLOGY

A. Data-set

We have evaluated the proposed action parsing algorithm on the Breakfast activity data-set [8]. The data-set includes a total of 52 participants, each performing a total of 10 cooking activities in multiple real-life kitchens. The data-set contains annotation of the type and the timing of the action units and complex cooking activities. Action units [8] comprise a pair of an action and an object involved in that action. Each instance of an action, such as pouring, stirring or cutting would be segmented as one action segment. Complex activities, such as making sandwich, making coffee or cooking egg, are composted of multiple actions. Each video in the dataset corresponds to one complex activity.

The Breakfast action recognition data-set contains annotations for 49 action units, 47 of them related to preparing foods for breakfast. Parsing the videos of cooking activities to action segments of action units is challenging. Even when action units are pre-segmented using ground truth annotation, estimating the action unit class of pre-segmented action segments has very low accuracy due to visual similarity of different action units and the fine granularity of movements involved in performing different food preparation actions.

Temporally segmenting a video of a cooking activity to action unit segments is additionally challenging, as the transitions between subsequent action units are often smooth and gradual.

B. Evaluated recognition model

The multi-class SVM used in the temporal context scoring function $\tilde{\mathcal{F}}$ of Eq. (5) uses the multi-class optimization proposed by Crammer and Singer [1] and uses bag-of-word encoding of sparse STIP features [9]. The temporal context features are generated at 4 scales using fixed segment lengths of $l$ frames, $l \in \{75, 150, 225, 300\}$.

The multi-class SVM used in scoring function $\mathcal{F}_{score}$ also uses the optimization proposed by Crammer and Singer [1]. The returned classification confidence score $\xi(u, l)$ is the margin between the winner and the runner up classes. We have set the minimum and the maximum segment lengths in the parsing search algorithm of Fig. [2] (i.e., $l_{min}$ and $l_{max}$) to 40 and 400 frames respectively. Context features are pooled over all segments before and after the segment of interest.

C. Evaluation measure

To evaluate the action parsing performance we have computed the per frame action unit classification accuracy.

TABLE I. PER-FRAME ACCURACY OF PARSING ACTION UNITS.

| Method | Per-frame accuracy of parsing action units |
|---|---|
| Sliding window, L = 75 | 0.146 |
| Sliding window, L = 150 | 0.179 |
| Sliding window, L = 225 | 0.187 |
| Sliding window, L = 300 | 0.189 |
| Hoai et al. [5] | 0.146 |
| Kuehne et al. [8] | 0.288 |
| **Proposed** | **0.301** |

## V. EVALUATION

### A. Experimental results

Table I shows per-frame accuracy of parsing videos in Breakfast activity data-set [8] to action units. Each frame of the video is classified to one of the 49 action unit classes. The accuracy is calculated by finding the ratio of the frames with the same estimated label as the ground truth annotation provided in the data-set. Note that some of the action unit classes such as 'pour milk', 'pour juice', 'pour water', 'pour sugar' and 'pour flour', are semantically and visually very similar. Considering the temporal context around these action units is expected to be strongly beneficial for distinguishing between these visually similar action unit classes.

Our proposed method achieved per-frame accuracy of 0.30. We have compared the accuracy of the proposed method with sliding window search of various sizes. As shown in Table I., the accuracy of sliding window search is only 0.19. We have also compared the accuracy of the proposed method with the joint segmentation and classification method of [5], which achieved an accuracy of 0.15. Note that the evaluation results for the method proposed in [5] is generated based on our re-implementation of the method described in the paper. In this evaluation, the minimum and the maximum candidate segment lengths (i.e., $l_{min}$ and $l_{max}$) are limited to 40 and 400 frames respectively.

Kuehne et al. has reported per-frame action unit parsing accuracy of 0.29 [8]. The parsing method proposed in [8] uses a high-level grammar which models the order the action units can appear in the sequence of different cooking activities. The grammar and the annotation of the cooking activity, which happens in each sequence, are the additional inputs used in the grammar based parsing method of [8].

### B. Analysis

We expect that the classification accuracy can be further improved by using the segments generated from the proposed parsing method of section III as input in a structured probabilistic graphical model such as CRF. The CRF models the temporal dependency of action unit classes of successive parsed segments, as well as the dependencies between the action unit class of segments and the class of the cooking activity that is performed in the sequence.

## VI. CONCLUSIONS

The segmentation of video sequences with an a priori unknown number of action segments can be improved through the use of temporal context information. We proposed a parsing algorithm which enables the use of temporal context information while searching for the optimal segmentation. Evaluation on the challenging Breakfast activity data-set [8] showed that the proposed method improves segmentation accuracy, even compared to a method based on high level grammar structure.

## ACKNOWLEDGMENT

This research was funded by Canon Inc. We thank them for their support. We also thank Xianghang Liu for his assistance with implementing the dynamic programming algorithm proposed by Hoai et al. [5].